%% file: main.tex
\pdfoutput=1

\documentclass{article}
\input{commands}

\begin{document}

\twocolumn[
\icmltitle{Evaluating Agents without Rewards}
\begin{icmlauthorlist}
\icmlauthor{Brendon Matusch}{vec}
\icmlauthor{Jimmy Ba}{vec,uoft}
\icmlauthor{Danijar Hafner}{vec,uoft,brain}
\end{icmlauthorlist}
\icmlaffiliation{vec}{Vector Institute}
\icmlaffiliation{uoft}{University of Toronto}
\icmlaffiliation{brain}{Google Brain}
\icmlcorrespondingauthor{Danijar Hafner}{mail@danijar.com}
\vskip 0.3in
]

\printAffiliationsAndNotice{}

\begin{abstract}
\begin{hyphenrules}{nohyphenation}
\input{sections/abstract}
\end{hyphenrules}
\end{abstract}

\input{sections/intro}
\input{sections/background}
\input{sections/method}
\input{sections/experiments}
\input{sections/discussion}

\clearpage
\begin{hyphenrules}{nohyphenation}
\bibliography{references}
\end{hyphenrules}
\clearpage
\appendix
\counterwithin{figure}{section}
\counterwithin{table}{section}
\input{sections/appendix}

\end{document}

%% file: commands.tex
\usepackage[accepted]{icml2021}
\usepackage[utf8]{inputenc}
\usepackage[T1]{fontenc}
\usepackage[english]{babel}
\usepackage{amsfonts}
\usepackage{nicefrac}
\usepackage{microtype}
\usepackage{amsmath}
\usepackage{amssymb}
\usepackage{mathtools}
\usepackage{subcaption}
\usepackage{booktabs}
\usepackage{natbib}
\usepackage{ragged2e}
\usepackage{siunitx}
\usepackage{tikz}
\usepackage{stackengine}
\usepackage{etoolbox}
\usepackage{xspace}
\usepackage{xpatch}
\usepackage{cuted}
\usepackage{enumerate}
\usepackage{xstring}
\usepackage{setspace}
\usepackage{tabularx}
\usepackage{makecell}
\usepackage{wrapfig}
\usepackage{graphicx}
\usepackage{changepage}
\usepackage{enumitem}
\usepackage{cuted}
\usepackage{titlesec}
\usepackage{cancel}
\usepackage{eqparbox}
\usepackage{multicol}
\usepackage{multirow}
\usepackage{bbm}
\usepackage{titlecaps}
\usepackage[export]{adjustbox}
\usepackage[hang,flushmargin]{footmisc}
\usepackage[useregional=numeric]{datetime2}
\usepackage[colorlinks]{hyperref}
\usepackage{url}
\usepackage[capitalise,noabbrev,nameinlink]{cleveref}

\usetikzlibrary{positioning,arrows.meta,fit,calc,backgrounds}

\setlist[itemize]{leftmargin=1.5em,itemsep=.05em,topsep=0em}
\titlespacing*{\section}{0pt}{1.2ex plus .5ex minus .5ex}{.3ex plus .1ex minus .1ex}
\titlespacing*{\subsection}{0pt}{.8ex plus .3ex minus .3ex}{.2ex plus .1ex minus .1ex}
\titlespacing*{\paragraph}{0pt}{.05ex plus .04ex minus .04ex}{1em}
\Addlcwords{and as but for if nor or so yet a an the as at by for in of off on per to up via}
\setcitestyle{authoryear,round,citesep={;},aysep={,},yysep={;}}
\renewcommand{\cite}[1]{\citep{#1}}

\creflabelformat{equation}{#2\textup{#1}#3}
\crefname{algocf}{Algorithm}{Algorithms}
\Crefname{algocf}{Algorithm}{Algorithms}

\definecolor{mydarkblue}{rgb}{0,0.08,0.45}
\hypersetup{colorlinks=true,linkcolor=mydarkblue,citecolor=mydarkblue,filecolor=mydarkblue,urlcolor=mydarkblue}

\graphicspath{{figures/}}

\makeatletter
\newcommand{\removeParBefore}{\ifvmode\vspace*{-\baselineskip}\setlength{\parskip}{0ex}\fi}
\newcommand{\removeParAfter}{\@ifnextchar\par\@gobble\relax}
\newcommand{\eq}{\begingroup\removeParBefore\endlinechar=32 \eqinner}
\newcommand{\eqinner}[2][aligned]{\endlinechar=32%
\begin{gather}\begin{#1}#2\end{#1}\end{gather}\endgroup\removeParAfter}
\makeatother

\DeclareDocumentCommand{\p}{ D<>{p} D<>{} D(){} }{\ensuremath{
\def\content{#3}\patchcmd{\content}{|}{\;#2\vert\;}{}{}
#1\ifdefempty{\content}{}{#2(\content #2)}}}

\DeclareDocumentCommand{\P}{ D<>{P} D<>{} D(){} }{\ensuremath{
\def\content{#3}\patchcmd{\content}{|}{\;#2\vert\;}{}{}
\operatorname{#1}\ifdefempty{\content}{}{#2(\content #2)}}}

\DeclareDocumentCommand{\E}{ D<>{E} E{_}{{}} D<>{\big} r[] }{
\def\content{#4}\patchcmd{\content}{|}{\;#3\vert\;}{}{}
\ensuremath{\operatorname{#1}_{#2}\!#3[\content #3]}}

\DeclareDocumentCommand{\D}{ D<>{D} D<>{\big} r[] }{
\def\content{#3}\patchcmd{\content}{||}{\;#2\|\;}{}{}
\ensuremath{\operatorname{#1}\!#2[\content #2]}}

\NewDocumentCommand{\Nor}{ r() }{\P<Normal>](#1)}
\NewDocumentCommand{\Cat}{ r() }{\P<Cat>](#1)}
\NewDocumentCommand{\Bin}{ r() }{\P<Bin>](#1)}
\NewDocumentCommand{\Bet}{ r() }{\P<Beta>](#1)}
\NewDocumentCommand{\Ber}{ r() }{\P<Bernoulli>(#1)}
\NewDocumentCommand{\Dir}{ r() }{\P<Dir>(#1)}

\DeclareDocumentCommand{\KL}{ D<>{\big} r[] }{\D<KL><#1>[#2]}
\DeclareDocumentCommand{\H}{ D<>{\big} E{_}{{}} r[] }{\E<H>_{#2}<#1>[#3]}
\DeclareDocumentCommand{\I}{ D<>{\big} E{_}{{}} r[] }{\E<I>_{#2}<#1>[#3]}

%% file: sections/abstract.tex
Reinforcement learning has enabled agents to solve challenging tasks in unknown environments. However, manually crafting reward functions can be time consuming, expensive, and error prone to human error. Competing objectives have been proposed for agents to learn without external supervision, but it has been unclear how well they reflect task rewards or human behavior. To accelerate the development of intrinsic objectives, we retrospectively compute potential objectives on pre-collected datasets of agent behavior, rather than optimizing them online, and compare them by analyzing their correlations. We study input entropy, information gain, and empowerment across seven agents, three Atari games, and the 3D game Minecraft. We find that all three intrinsic objectives correlate more strongly with a human behavior similarity metric than with task reward. Moreover, input entropy and information gain correlate more strongly with human similarity than task reward does, suggesting the use of intrinsic objectives for designing agents that behave similarly to human players.

%% file: sections/intro.tex
\vspace*{-3ex}
\section{Introduction}
\label{sec:intro}

Reinforcement learning (RL) has enabled agents to solve complex tasks directly from high-dimensional image inputs, such as locomotion \citep{heess2017parkour}, robotic manipulation \citep{akkaya2019rubiks}, and game playing \citep{mnih2015dqn,silver2017alphago}. However, many of these successes are built upon rich supervision in the form of manually defined reward functions. Unfortunately, designing informative reward functions is often expensive, time-consuming, and prone to human error \citep{krakovna2020specification}. Furthermore, these difficulties increase with the complexity of the task of interest.

In contrast to many RL agents, natural agents frequently learn without externally provided tasks, through intrinsic objectives. For example, children explore the world by crawling around and playing with objects they find. Inspired by this, the field of intrinsic motivation \citep{schmidhuber1991curiousmodel,oudeyer2007curiosity} seeks mathematical objectives for RL agents that do not depend on a specific task and can be applicable to any unknown environment. We study 3 common types of intrinsic motivation:

\input{tables/first}
\input{figures/method/figure}
\input{figures/envs/figure}

\begin{itemize}
\item Input entropy encourages encountering rare sensory inputs, measured by a learned density model \citep{schmidhuber1990diffmodel,bellemare2016pseudocount,pathak2017icm,burda2018rnd}.
\item Information gain rewards the agent for discovering the rules of its environment \citep{lindley1956expectedinfo,houthooft2016vime,shyam2018max,sekar2020plan2explore}.
\item Empowerment rewards the agent's influence over its sensory inputs or the environment \citep{klyubin2005empowerment,mohamed2015empowerment,karl2017empowerment}.
\end{itemize}

Despite the empirical success of intrinsic motivation for facilitating exploration \citep{bellemare2016pseudocount,burda2018rnd}, it remains unclear which family of intrinsic objectives is best for a given scenario, for example when task rewards are sparse or unavailable, or when the goal is to behave similarly to humans. Designing intrinsic objectives that result in intelligent behavior across different environments is an important unsolved problem. Moreover, it is not clear whether different intrinsic objectives offer similar benefits in practice or are orthogonal. Progress toward answering these questions is hindered by slow iteration speeds because for each new intrinsic objective, one typically needs to design and train a new agent in one or more environments.

To address these challenges, we propose the methodology of evaluating and comparing intrinsic objectives by correlation analysis on a fixed dataset, as shown in \cref{fig:method}. This frees us from having to train a separate agent for every objective considered, and alleviates the complexity associated with optimizing objectives online. We additionally compare a human similarity measure that is computed based on recorded trajectories of human players. To conduct the study, we collect a diverse dataset of 26 agents in 4 complex environments to compare task reward, human similarity, and 3 intrinsic objectives. The dataset contains a total of over 2 billion images and is made available freely. The key findings of the experimental study are summarized as follows:
\begin{itemize}
\item Input entropy and information gain correlate more strongly with human similarity than task reward does. The intrinsic objectives also correlate strongly with human similarity across all studied environments, while reward does not. To develop agents that behave similarly to human players, intrinsic objectives may thus be more relevant than typical task rewards.
\item Simple implementations of input entropy, information gain, and empowerment based on image discretization correlate strongly with human similarity. Thus, they could lead to effective exploration when optimized online and could serve as evaluation metrics when task rewards and demonstrations are unavailable.
\item Input entropy and information gain correlate strongly with each other, but to a lesser degree with empowerment. This suggests that optimizing empowerment together with either of the two other objectives could be beneficial for designing exploration methods.
\end{itemize}

The methodology of retrospectively comparing agent objectives based on their correlations has the potential to accelerate future research on developing exploration objectives in reinforcement learning and for finding objectives that capture aspects of human behavior in the cognitive sciences.

%% file: tables/first.tex
\begin{table}[tb!]
\centering
\vspace*{5.1ex}
\begin{tabularx}{\linewidth}{Xcc}
\toprule
\textbf{Objective} &
\makecell{\textbf{Reward} \\ \textbf{Correlation}} &
\makecell{\textbf{Human Similarity} \\ \textbf{Correlation}} \\
\midrule
Task Reward & \textbf{1.00} & 0.67 \\
Human Similarity & 0.67 & \textbf{1.00} \\
Input Entropy & 0.54 & \textbf{0.89} \\
Information Gain & 0.49 & \textbf{0.79} \\
Empowerment & 0.41 & \textbf{0.66} \\
\bottomrule
\end{tabularx}
\caption{We computed Pearson correlation coefficients of each intrinsic objective with task reward and human similarity across 3 Atari games and Minecraft from over 2 billion time steps. The intrinsic objectives correlate more strongly with human similarity than with task reward. This suggests that typical RL tasks may not be a sufficient proxy for the behavior that is seen in humans playing the same games.}
\label{tab:first}
\end{table}

%% file: figures/method/figure.tex
\begin{figure*}[t]
\centering
\includegraphics[width=0.93\textwidth]{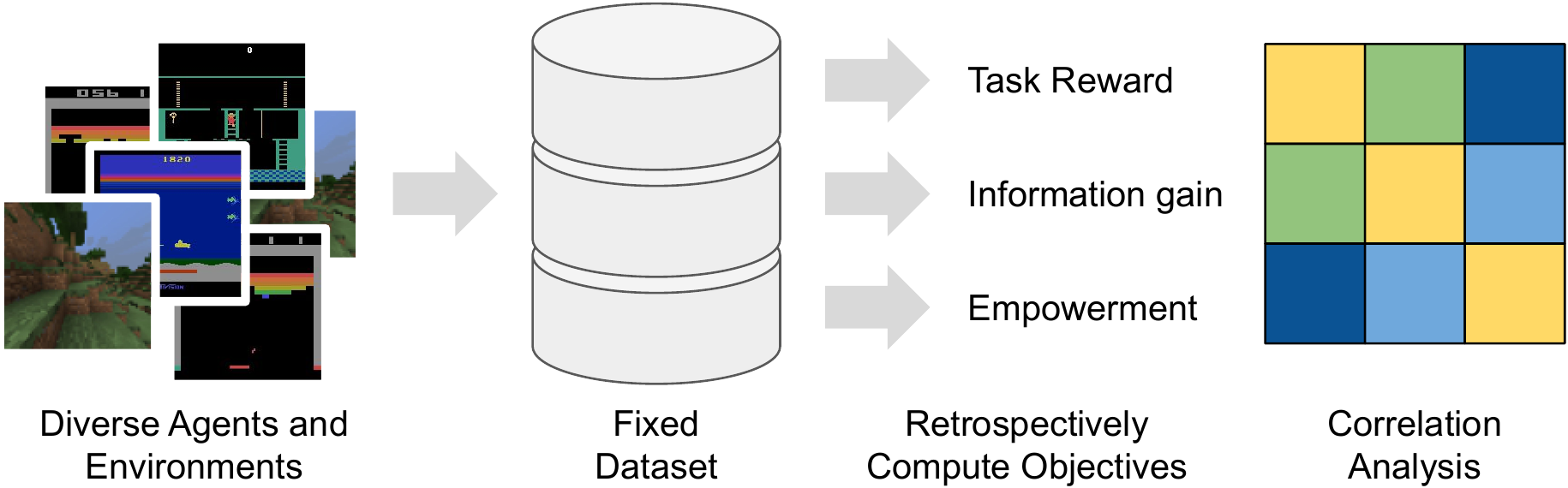}
\caption{Training agents to evaluate different intrinsic objectives can be a slow and expensive process. To address this problem, we collect a diverse dataset of different environments and behaviors once and retrospectively compute agent objectives from it. We then analyze the correlations between intrinsic objectives and supervised objectives, such as task reward and human similarity. This speeds up the iteration time by letting us draw conclusions on the relationships between different intrinsic objectives without having to train a new agent for each of them.}
\label{fig:method}
\vspace*{-1ex}
\end{figure*}

%% file: figures/envs/figure.tex
\newcommand{\grayarrow}{%
\rotatebox{270}{%
\begin{tikzpicture}[x=2em,y=2em,baseline=1em]
\clip (0,0) rectangle + (1,1);
\draw[-{Triangle[length=1.4em,width=2em]},line width=1em,gray] (0,0.5) -- (1,0.5);
\end{tikzpicture}}}
\begin{figure*}[t]
\centering
\newcolumntype{Y}{>{\centering\arraybackslash}X}
\begin{tabularx}{\textwidth}{YYYY}
Breakout & Seaquest & Montezuma & Minecraft \\[1ex]
\includegraphics[width=0.7\linewidth,height=0.7\linewidth]{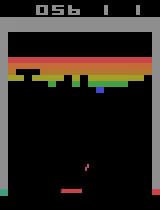} &
\includegraphics[width=0.7\linewidth,height=0.7\linewidth]{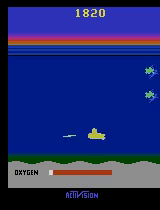} &
\includegraphics[width=0.7\linewidth,height=0.7\linewidth]{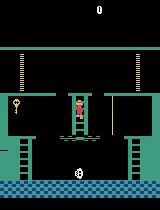} &
\includegraphics[width=0.7\linewidth,height=0.7\linewidth]{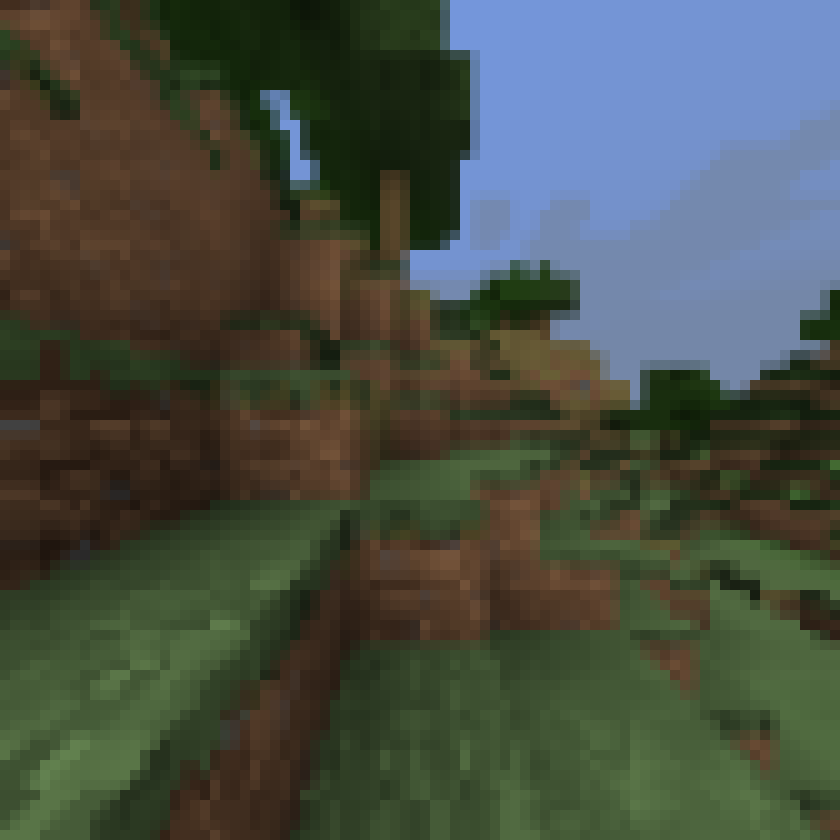} \\[-1ex]
\grayarrow & \grayarrow & \grayarrow & \grayarrow \\[5ex]
\frame{\includegraphics[width=0.7\linewidth]{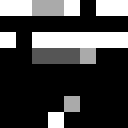}} &
\frame{\includegraphics[width=0.7\linewidth]{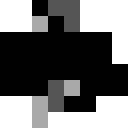}} &
\frame{\includegraphics[width=0.7\linewidth]{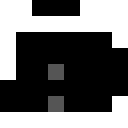}} &
\frame{\includegraphics[width=0.7\linewidth]{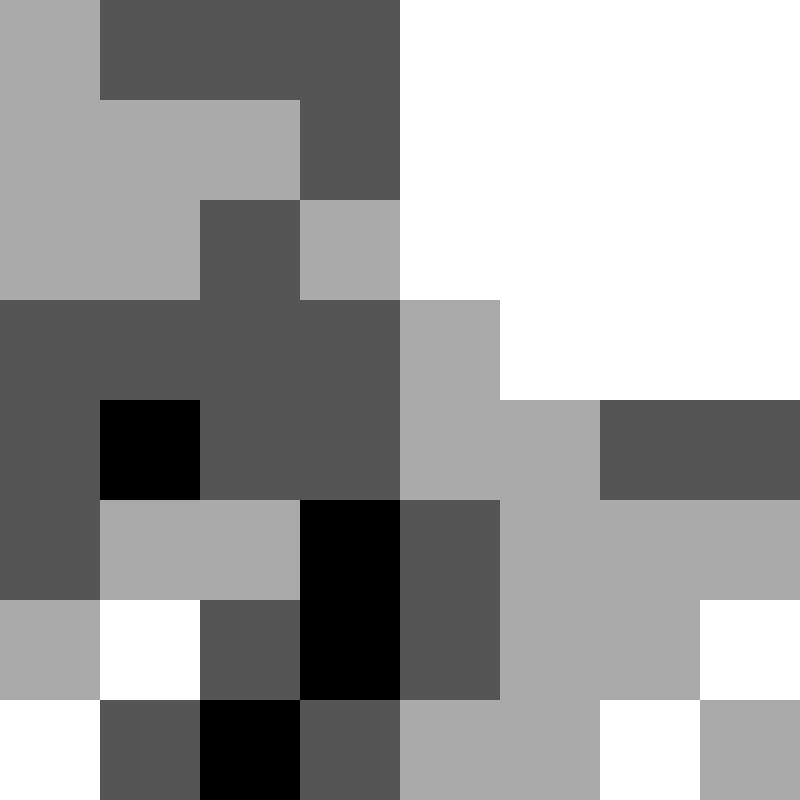}} \\
\end{tabularx}
\caption{To make computing the agent objectives tractable and efficient, we preprocess the images in our collected datasets by discretizing them into buckets. Similar to Go-Explore \citep{ecoffet2019goexplore}, we resize the images to $8 \times 8$ pixels and discretize each of the resulting cells to one of $4$ values. The examples show that this procedure preserves positions of objects in the game, such as the player, ball, fish, and skull. We enumerate the discretized images to represent each unique frame by an integer index to compute discrete probability tensors for the environments.}
\label{fig:envs}
\vspace*{-1ex}
\end{figure*}

%% file: sections/background.tex
\section{Background}
\label{sec:background}

To validate the effectiveness of our intrinsic objectives across a wide spectrum of agent behavior, we retrospectively computed our objectives on the lifetime experience of well-known RL agents. Thus, we first collected datasets of various agent behaviors on which to compare our objectives.

\paragraph{Environments}

We chose three different Atari environments provided by Arcade Learning Environment \citep{bellemare2013ale}: Breakout, Seaquest, and Montezuma's Revenge, and additionally the Minecraft Treechop environment provided by MineRL \citep{guss2019minerldata}. Breakout and Seaquest are relatively simple reactive environments, while Montezuma is a challenging platformer requiring long-term planning. Treechop is a 3D environment in which the agent receives reward for breaking and collecting wood blocks, but has considerable freedom to explore the world. As detailed in \cref{sec:envs}, we chose these four environments because they span a range of complexity, freedom, and difficulty.

\paragraph{Agents}

The 7 agents represented in our dataset include 3 learning algorithms and 2 trivial agents for comparison. We selected RL agents spanning the range from extrinsic task reward to intrinsic motivation reward. Additionally, we included random and no-op agents, two opposite extremes of naive behavior. Our goal was to represent a wide range of behaviors: playing to achieve a high score, playing to explore the environment, and taking actions without regard to the environment. Specifically, we used the PPO agent \citep{schulman2017ppo} trained to optimize task reward, and the RND \citep{burda2018rnd} and ICM \citep{pathak2017icm} exploration agents that use PPO for policy optimization. We train the exploration agents, with default hyperparameters, once using only the intrinsic reward and once using both intrinsic and task rewards. The agents are described further in \cref{sec:agents}.

%% file: sections/method.tex
\section{Method}
\label{sec:method}

To spur progress toward better understanding of intrinsic objectives, we empirically compared the three objective families in terms of their correlation with human behavior and with the task rewards of three Atari games and Minecraft. First, we trained several well-known RL agents on three Atari games and Minecraft and store their lifetime datasets of experience. Specifically, 100 million frames on each of the three Atari environments with each of seven agents: random, no-op, PPO, and RND and ICM with and without task reward. Minecraft was evaluated for 12 million frames per agent because the simulation is slower than the Atari games, and five agents rather than seven were used, excluding both configurations of ICM. This resulted in a total of 2.1 billion time steps and about 9 terabytes of agent experience.

\raggedbottom\pagebreak

We preprocessed the experience datasets and computed human similarity, input entropy, empowerment, and information gain using simple estimators with clearly stated assumptions, in aggregate over each agent lifetime, yielding one number per objective-agent-environment. A table of all computed values is included in \cref{tab:metrics}. We then analyzed the correlations between the intrinsic objectives to understand how they relate to another and how well they reflect task reward and human similarity. We now describe the preprocessing and introduce our estimators for the objectives.

\subsection{Preprocessing}

To make the computation of the considered objectives tractable and efficient, we discretize the agent's input images so that they can be represented by bucket indices. This allows us to summarize each collected dataset as a sparse tensor that holds the counts of each possible transition tuple, from which we then compute the objective values.

\paragraph{Discretization}
As shown in \Cref{fig:envs}, we preprocess the images by first converting them from RGB to grayscale as they were seen by the agents. After that, we bilinearly resize them to $8 \times 8$ pixels. We discretize these low-resolution images to four possible values per pixel, with thresholds chosen as the brightness percentiles 25, 50, and 75 across all unique values of the corresponding pixel in the environment across all agents. We also considered choosing the thresholds based on each agent individually, as discussed in \cref{sec:infogain_variants}. The unique discretized images are enumerated to represent each image by an integer index.

\paragraph{Aggregation}
For each pairing of agent and environment, we summarize the transitions preprocessed images and actions into a tensor of counts. For an image index
$1 \leq i \leq |X|$, and action index $1 \leq j \leq |A|$, and a successor image index $1 \leq k \leq |X|$, where $X$ is the set of inputs and $A$ the set of actions, the count tensor $N_{ijk}$ is defined as the number of transitions from image bucket $i$ and action $j$ to image bucket $k$. We simply chose the discretization parameters ($8 \times 8$ pixels with $4$ values per pixel) to make the objectives computationally feasible and did not tune them further, suggesting that this approach is not sensitive to these values.

Normalizing the count tensor $N$ yields a probability tensor $P$ that stores the probability of each transition in the agent's dataset. Under the assumption of a Markovian environment and agent, the probability tensor fully describes the statistics of the preprocessed dataset,

\eq{P \doteq N \Big/ \sum_{ijk} N_{ijk}.}

The probability tensor $P$ describes the joint probability of transitions for each agent and environment and thus allows us to compute any marginals and conditionals needed for computing the objectives.

\subsection{Objectives}

We compare two supervised objectives, task reward and human similarity, as well as three intrinsic objectives: input entropy, information gain, and empowerment. We compute a single value of each of these objectives on each agent-environment dataset.

\paragraph{Task reward}
The reward provided by RL environments measures success at a specific task. The environments we use have only one predefined task each, despite the wide range of conceivable objectives, especially in Montezuma's Revenge and Minecraft. This is true of many RL environments, and limits one's ability to analyze the behavior of an agent in a general sense. Multi-task benchmarks address this problem but often include a distinct environment for each task rather than multiple tasks within the same environment \citep{yu2019metaworld}. This would make it difficult to evaluate the agent's ability to globally explore its environment independent of the task.

\paragraph{Human similarity}
Task reward captures only the agent's success at the specific task defined via the reward function. This may not match up with the behavior of a human player who interacts with an environment. To capture this human-like aspect of behavior, we compute the similarity between the agent's behavior and human behavior in the same environment, that is, using human behavior as a ``ground truth.'' Inspired by the inverse RL literature \citep{ziebart2008maximum, klein2012inverse}, we measure the overlap between human and agent observations. We leverage the Atari-HEAD \citep{zhang2019atarihead} and the MineRL Treechop \citep{guss2019minerldata} datasets and preprocess them the same way as the agent datasets. The Atari human datasets contain $\sim\!250\mathrm{K}$ frames per environment, and $\sim\!460\mathrm{K}$ frames are available for Minecraft.

We compute human similarity as the Jaccard index, also known as intersection over union, between the unique input images encountered in the human dataset and those encountered by the artificial agent. For this, we first compute the marginal input probabilities from the probability tensors $\smash{P^{\text{\,agent}}}$ and $\smash{P^{\text{\,human}}}$ of the artificial agent and the human player, respectively. The human similarity is then computed as a fraction of non-zero probability entries, where $\operatorname{\mathbbm{1}}(\cdot)$ denotes the indicator function that evaluates to $1$ for true inputs and to $0$ for false inputs,

\eq{
\begin{gathered}
\operatorname{S} \doteq
\sum_i \operatorname{\mathbbm{1}}(X_i \!>\! 0 \land Y_i \!>\! 0) \Big/
\sum_i \operatorname{\mathbbm{1}}(X_i \!>\! 0 \lor Y_i \!>\! 0) \\
X_i \doteq \sum_{jk} P^{\text{\,agent}}_{ijk}, \quad
Y_i \doteq \sum_{jk} P^{\text{\,human}}_{ijk}.
\end{gathered}
\raisetag{7ex}
}

Note that while we use input images from recorded human behavior to compute human similarity, we do not compare the human and agent actions directly, as the RL agents play in an environment with sticky actions, while the humans did not. Collecting human datasets is challenging in environments that are challenging for human players, such as high-dimensional continuous control. Thus, we consider three intrinsic objectives, which do not require environment-specific engineering or human demonstrators.

\input{figures/scores/figure}

\paragraph{Input entropy}
The agent's input entropy in the environment measures how spread out its visitation distribution over inputs is. The input entropy tends to be larger the more inputs the agent has reached, and the more uniformly it visits them. In other words, input entropy measures how improbable individual inputs are under the input distribution of the agent. This idea has been used for exploration in RL, where the negative log-probability of inputs under a learned density model is used as exploration bonus \citep{schmidhuber1991curiousmodel,oudeyer2007curiosity, bellemare2016cts,burda2018rnd}. Because we compute the input entropy retrospectively, we have access to the agent's input distribution via the probability tensor that summarizes the agent's lifetime experience. We compute the the entropy over inputs $x$ by marginalizing out actions and success inputs,

\eq{\operatorname{C} \doteq \H[x] = - \sum_{i} X_i \ln X_i, \quad X_i \doteq \sum_{jk} P_{ijk}.}
\vspace*{-3ex}

\paragraph{Information gain}
Information gain measures how many bits of information the agent has learned about the environment from its dataset of experience \citep{lindley1956expectedinfo}. It is the mutual information between observations and the agent's representation of the environment. Information gain has led to successful exploration in RL \citep{sun2011plansurprise, houthooft2016vime,shyam2018max,mirchev2018dvbflm,sekar2020plan2explore}.

To measure the amount of information gained, we need a way to choose a representation of the environment that summarizes the agent's knowledge. Preprocessing the inputs into discrete classes enables us to represent the knowledge as a belief over the true transition matrix. The total information gain of over agent's lifetime is the entropy difference of the prior belief over the transition matrix and the posterior belief. We choose the beliefs to be Dirichlet distributions over successor inputs $x'$  for every pairing of current inputs $x$ and actions $a$ \citep{sun2011plansurprise,friston2017actinfparaminfogain},

\eq{
\operatorname{I} &\doteq \I[z; d] = \H[z] - \H[z | d] \\[1ex]
&= \sum_{ij} \H[\Dir(1)] - \sum_{ij} \H[\Dir(1+N_{ij})].
}

The Dirichlet distribution is parameterized by a vector of counts, known as concentration parameter $\alpha$. There is one Dirichlet distribution for each pairing of current input $x$ and action $a$, with a concentration parameter of the length of possible successor inputs $x'$. We choose a uniform Dirichlet distribution as prior belief, corresponding to a vector of ones for its concentration parameter. The posterior belief is typically a Dirichlet belief that uses as concentration parameter the prior value of $1$ plus the count vector $N_{ij}$.

\input{figures/bars/figure}


The entropy of a Dirichlet distribution is computed from the vector of concentration parameters $\alpha$ using the digamma function $\operatorname{\psi}(\cdot)$ the incomplete beta function $\operatorname{B}(\cdot)$ \citep{lin2016dirichlet},

\eq{
\H[\Dir(\alpha)] &=
\ln \operatorname{B}(\alpha)
- \sum_k (\alpha_k - 1) \operatorname{\psi}(\alpha_k) \\
&+ \Big(\sum_k \alpha_k - |X|\Big) \operatorname{\psi}\Big(\sum_k \alpha_k\Big).
}

In practice, we find it more effective to only consider unique transitions, and thus we use $1+\operatorname{sign}(N_{ij})$ as the posterior concentration instead of the prior plus raw counts $1+N_{ij}$. We conjecture that choice leads to a more meaningful information gain objective because the four environments are close to deterministic, with over 80\% of input action pairs leading to a unique successor input in the datasets. Thus, experiencing a transition once establishes that the transition is possible, and revisiting it multiple times should not further increase the agent's knowledge. We compare alternatives to this choice in \cref{sec:infogain_variants}.

\paragraph{Empowerment}
Empowerment measures the agent's influence over its sensory inputs and thus the environment \citep{klyubin2005empowerment}. It has been applied to RL by \citet{mohamed2015empowerment,salge2013continuousempow,karl2017empowerment,leibfried2019empowreward,zhao2020empow}. Multiple variants exist in the literature, including potential empowerment where the agent aims to ``have many options'' and realized empowerment where the agent aims to ``use many options.'' We consider the latter variant, that is measured as measured as the mutual information between the agents actions and resulting inputs, given current inputs \citep{salge2014realizedempow,hafner2020apd}. We compute the one-step empowerment as the difference between the entropy of action $a$ given the preceding input $x$, before and after observing the following input $x'$,

\eq{
\begin{gathered}
\begin{aligned}
\operatorname{E} &\doteq \I[x'; a | x] = \H[a | x] - \H[a | x, x'] \\[1ex]
&= \sum_{ijk} P_{ijk} \ln X_{ijk} - \sum_{ijk} P_{ijk} \ln Y_{ij}, \\
\end{aligned} \\
X_{ijk} = P_{ijk} \Big/ \sum_{i'k'} P_{i'jk'}, \quad
Y_{ij}  = \sum_{k'} P_{ijk'} \Big/ \sum_{i'k'} P_{i'jk'}.
\end{gathered}
\raisetag{10ex}
}

%% file: figures/scores/figure.tex
\begin{figure*}[t]
\centering
\includegraphics[width=0.92\textwidth]{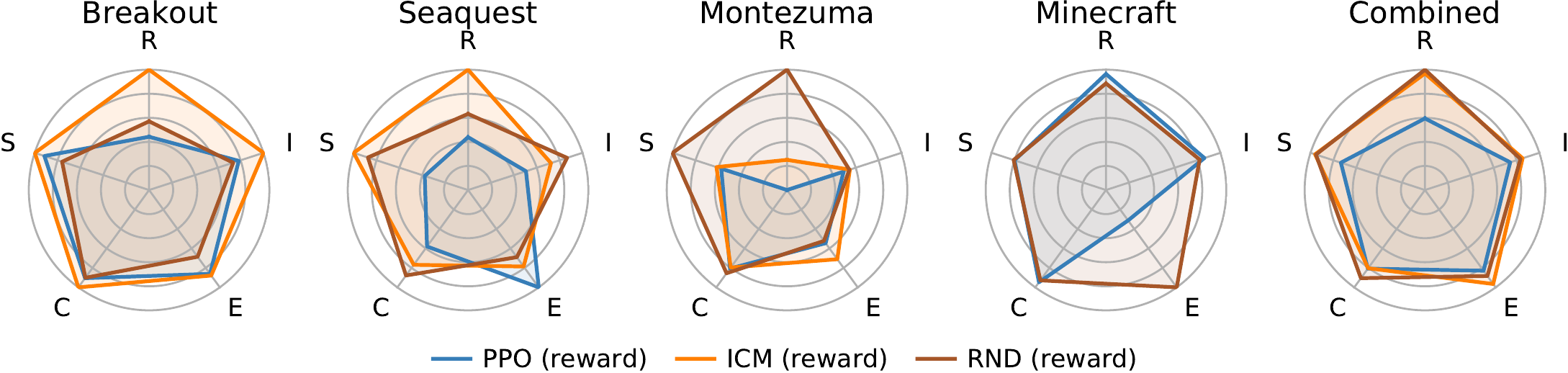}
\vspace*{1ex} \\
\includegraphics[width=0.92\textwidth]{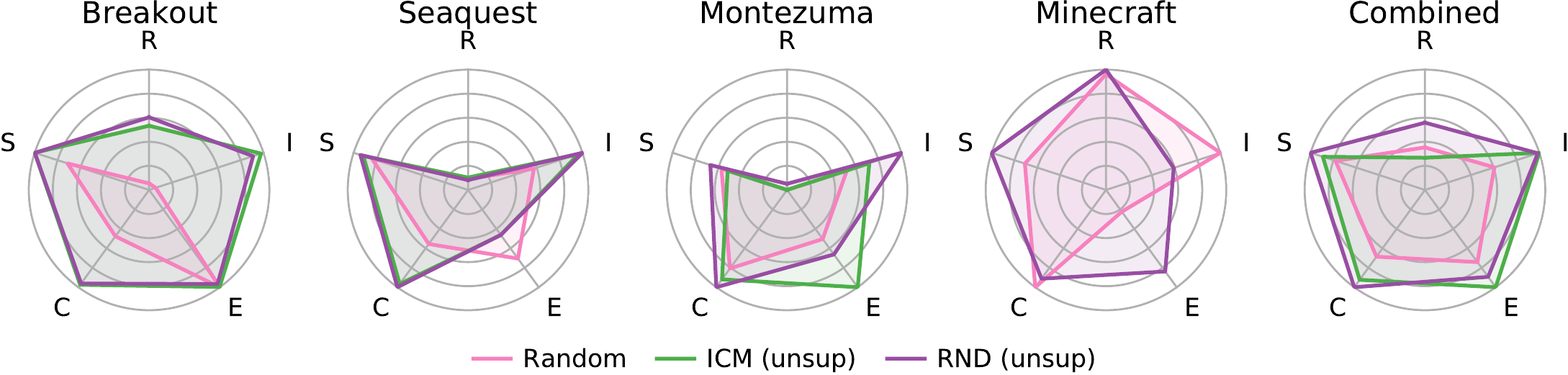}
\caption{Objective values for agents that use task reward (top) and agents without access to task reward (bottom). The supervised objectives are task reward (R) and human similarity (S) and the intrinsic objectives are input entropy (C), empowerment (E), and information gain (I). The no-op agent achieves the lowest scores in all objectives does not show up in the normalized coordinates. The two exploration agents with access to task rewards achieve the highest task reward and human similarity across Atari environments, and RND without reward in Minecraft. ICM or RND each achieve the highest input entropy and information gain value in three out of four environments according to our objectives. Surprisingly, PPO and task-agnostic ICM achieve high empowerment, even in Montezuma where they achieve low task reward. }
\label{fig:scores}
\end{figure*}

%% file: figures/bars/figure.tex
\begin{figure*}[th!]
\centering
\includegraphics[width=0.78\linewidth]{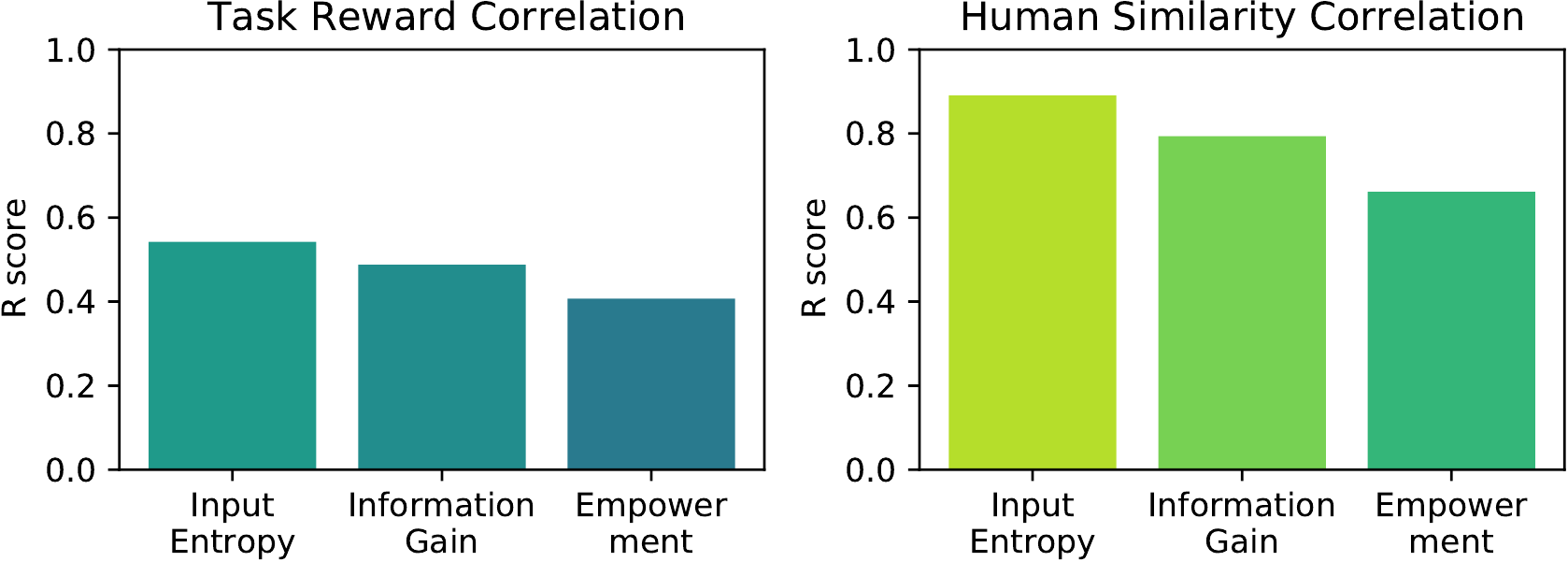}
\caption{Pearson correlation coefficients between the three considered intrinsic objectives and task reward (left) and similarity to human players (right). All considered intrinsic objectives correlate more strongly with human similarity than with the task rewards in the RL environments. This suggests that typical task rewards may be insufficient for evaluating exploration objectives when the goal is to produce behavior similar to human players. Interestingly, ranking the intrinsic objectives by their correlations with either task reward or human similarity gives the same ordering. The correlations were computed from $N=26$ agent environment pairings and are statistically significant with $p<0.05$.}
\label{fig:bars}
\end{figure*}

%% file: sections/experiments.tex
\section{Analysis}
\label{sec:experiments}

We conduct a correlation analysis to understand how the three intrinsic objectives relate to another and to the supervised objectives, task reward and human similarity. First, we compute all objectives for the agents and compare the agents by them. Second, we rank the intrinsic objectives by their correlations to task reward and human similarity. Third, we compare the correlations among the intrinsic objectives. We make the source code for replicating our analyses and the collected datasets available to support future work on evaluating agents without task rewards.\footnote{\url{https://danijar.com/agenteval}}

\subsection{Evaluation of Agents}

We begin by computing and analyzing the objective values for individual agents, with the goal of intuitively understanding what specific agent behavior leads to high and low values of each objective. RND and ICM have the highest values of task reward and human similarity in all environments; that the agent with highest input entropy and is also the agent with highest information gain in all environments; and that empowerment varies widely by environment. The objective values are visualized in \cref{fig:scores} and their numbers are given in \cref{sec:metrics}.

\paragraph{Task reward}
The total episode rewards of the task-specific RND and ICM agents we trained are comparable to those reported by \citet{taiga2020bonus}. ICM achieve slightly higher reward than RND in Seaquest, while the opposite is true in \citet{taiga2020bonus}. We observe in \cref{sec:metrics} that task-specific RND and ICM achieve the highest task reward per time step in the environments except in Minecraft, where task-agnostic RND performs best, showing that explicit exploration is beneficial in all our considered environments.

\paragraph{Human similarity}
Human similarity is the highest for task-specific ICM and RND in Seaquest and Montezuma respectively, but for task-agnostic ICM in Breakout and task-agnostic RND in Minecraft. Exploration agents achieve the highest human similarity in all four environments. In environments other than Minecraft, the random agent has substantially lower human similarity than the other agents, and no-op is consistently the lowest across all environments.

\input{figures/corr/figure}

\paragraph{Input entropy}
Task-agnostic ICM and RND obtain the highest input entropy in all environments, except in Minecraft where the random agent achieves the highest input entropy. This suggests that using task reward ``distracts'' the task-specific agents from maximizing input entropy. The no-op agent has the lowest input entropy in all environments. The random agent achieves high input entropy in Minecraft, where many distinct inputs are easy to reach from the initial state, e.g. by moving the camera.

\paragraph{Information gain}
Information gain is highest in Breakout for ICM with reward, in Seaquest and Montezuma for RND without reward, and in Minecraft for the random agent. We conjecture that the random agent would achieve lower information gain in Minecraft under a preprocessing scheme that groups semantically similar observations into the same bucket. In all four environments, the agent achieving the highest input entropy also achieves the highest information gain, implying that input entropy and information gain are closely related in practice.

\paragraph{Empowerment}
The agents that have the most empowerment vary more across environments than they do for the other objectives. In Seaquest, where most meaningful actions are tied to reward, the PPO agent achieves highest empowerment. In Montezuma, where many actions influence the input in ways unrelated to the task, task-agnostic ICM achieves the highest empowerment. In Breakout, where almost all actions move the paddle and thus influence the input, the random agent achieves the highest empowerment.

\subsection{Evaluation of Intrinsic Objectives}

We evaluate the three intrinsic objectives based on their correlations with task reward and human similarity. The correlations aggregated across all environments are shown in \cref{fig:bars} and their numeric values are included in \cref{tab:corr}. This constitutes our main result. All intrinsic objectives correlate more strongly with human similarity than with task reward. Moreover, ranking them by their correlations with either task reward or human similarity yields the same order. Finally, the intrinsic objectives correlate positively with task reward and human similarity within individual environments in almost all cases.

\paragraph{By task reward and human similarity}
All intrinsic objectives correlate substantially more with human similarity than with task reward. The correlations with human similarity are: input entropy (R=0.89), information gain (R=0.79), and empowerment (R=0.66). The correlations with task reward are: input entropy (R=0.54), information gain (R=0.49), and empowerment (R=0.41). This suggests that intrinsic objectives may capture more aspects of human behavior than typical task rewards, and may thus be the more promising approach when the goal is to design agents that behave similarly to humans.

\paragraph{Ranking of intrinsic objectives}
Ranking the intrinsic objectives by their correlations with either task reward or human similarity yields in the same order. Input entropy correlates most strongly with task reward and human similarity, followed by information gain, and then empowerment. This suggests that exploration objectives such as input entropy and information gain are beneficial across the four environments, whereas task reward, which is not directly encouraging exploration, does not seem as important. Moreover, our results identify input entropy as a promising intrinsic objective that should work well across many scenarios.

\newcommand{\tleq}{\!\ensuremath{\leq}}

\paragraph{Per-environment correlations}
We additionally include per-environment correlations in \cref{fig:corr}. Not only aggregated across environments, but also within each environment, the intrinsic objectives correlate positively with both task reward and human similarity in 23 out of 24 cases.
The correlations between intrinsic objectives and human similarity are strong in all four environments (0.56 \tleq R \tleq 0.96), except for empowerment in Seaquest (R=0.44). In contrast, the correlations with task reward vary more across environments, being larger in Breakout (0.48 \tleq R \tleq 0.91) and Minecraft (0.59 \tleq R \tleq 0.98) but weaker in Seaquest (0.22 \tleq R \tleq 0.61) and Montezuma (0.00 \tleq R \tleq 0.19).

\subsection{Comparison among Objectives}

The right-most correlation matrix in \cref{fig:corr} shows the correlations among all five objectives, which are statistically significant with $p<0.05$. The numerical values are included in \cref{tab:corr}. Task reward and human similarity correlate positively (R=0.67) but this correlation is weaker than those between human similarity and input entropy (R=0.89) or information gain (R=0.79). Input entropy and information gain correlate strongly (R=0.95), suggesting that they would have similar effects when optimized online. In contrast, empowerment correlates less strongly with the two objectives (R=0.66 and R=0.55). This suggests that empowerment measures a different component of behavior and combining it with either input entropy or information gain could be beneficial when designing novel exploration methods.

%% file: figures/corr/figure.tex
\begin{figure*}[t]
\includegraphics[width=\textwidth]{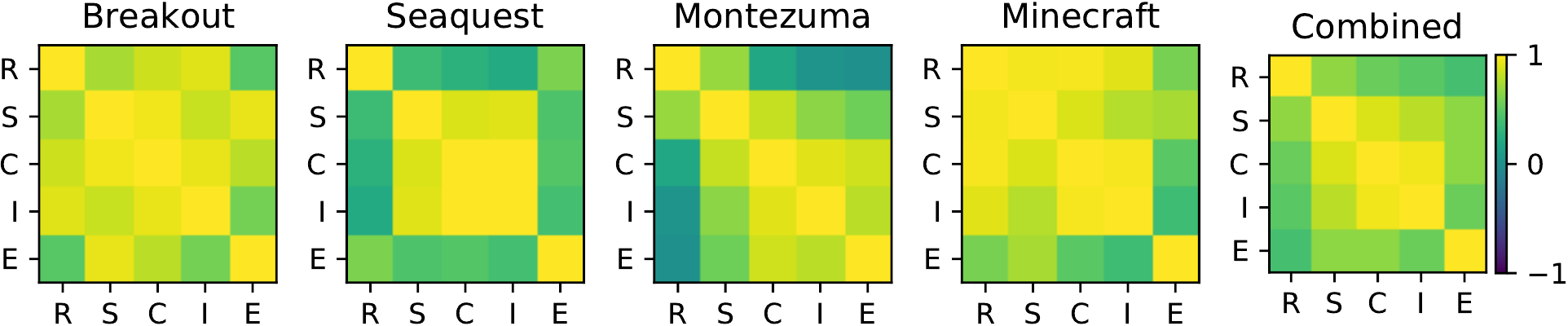}
\caption{Correlation matrices for the objectives in our study: task reward (R), human similarity (S), input entropy (C), empowerment (E), information gain (I). The correlation is taken across agents. All objectives correlate positively but to different degrees. While human similarity and the three intrinsic objectives all correlate strongly with another, task reward only correlates weakly with them on Seaquest and Montezuma. As per-environment correlations are computed from 7 agents in Atari environments and 5 in Minecraft, they are not statistically significant and provided only for exploratory analysis. The numerical values for the correlation matrices are included in \cref{tab:corr}.}
\label{fig:corr}
\vspace*{-1ex}
\end{figure*}

%% file: sections/discussion.tex
\section{Related Work}

Intrinsic objectives are often used as an exploration bonus that is combined with a task reward during training \citep{bellemare2016cts,houthooft2016vime,riedmiller2018sparseplaying}. In this case, the agent behavior is typically evaluated by task performance, which may be sparse rewards. \citet{pathak2018largescale} evaluate the ICM agent without extrinsic rewards in 54 environments and find successful performance as measured by coincidental task reward. Recently, \citet{taiga2020bonus} compare four exploration bonuses including ICM and RND on Atari games, finding that their benefits tend to vary across games.

In contrast, evaluating agents that learn without specific tasks in mind is more difficult and an open research challenge. Several approaches have been explored:

\begin{itemize}
\item Solving downstream tasks with a new agent that has access to the experience collected by the unsupervised agent \citep{lynch2019play,sekar2020plan2explore} requires domain knowledge for designing the downstream tasks.
\item Qualitative visualizations of agent behavior or skills \citep{gregor2016vic, eysenbach2018diayn,sharma2019dads,xie2020lsp}, which in isolation do not allow for quantitative comparison.
\item Environment-specific proxy metrics, such as the number of visited rooms \citet{bellemare2016pseudocount} or amount of map coverage \citep{shyam2018max} are easy to interpret but cannot be used in unknown environments.
\item General proxy metrics, such as complexity measures \citep{martius2015agentcomplexity}, empowerment \citep{gregor2016vic}, and information gain \citep{sekar2020plan2explore}. These metrics are generally applicable but it can be unclear what qualitative aspects of agent behavior they measure, and how they are related to each other in practice.
\item Co-incidental task reward, such as playing the game Mario purely by curiosity \citep{pathak2017icm}. This metric works for linear environments such as Mario but may fail when there are many possible tasks.
\end{itemize}

The connections between these different evaluation metrics are not yet well understood. Our paper takes a step toward developing this understanding through a correlation study across 4 environments with high-dimensional inputs and 7 agent variants, with focus on 3 general agent objectives: input entropy, information gain, and empowerment. We contribute to the interpretability of these 3 metrics by showing that the first two are closely related to each other and that all 3 correlate better with human similarity than with empowerment. Moreover, our retrospective analysis offers a simple framework for evaluating novel objectives for exploration.

Recently, \citet{kosoy2020exploring} compare children and RL agents that navigate a 3D maze using a free exploration phase followed by the downstream task of finding a goal. We formalize the notion of human similarity as a concrete metric, which could be beneficial for this line of work.




\section{Discussion}
\label{sec:discussion}

In this paper, we have collected large and diverse datasets of agent behavior, computed 3 intrinsic objectives on each dataset, and analyzed the correlations of the intrinsic objectives with each other and with task reward and a human similarity objective. Our retrospective evaluation methodology enabled us to compare different intrinsic objectives while avoiding the complexity and typically slow iteration speed that come with online training.

\paragraph{Key findings}
All studied intrinsic objectives correlate more strongly with human similarity than the task rewards do. These correlations hold consistently across all environments, whereas task reward and human similarity correlate strongly only in half of the environments. We thus recommend intrinsic objectives over task rewards when the goal is to design general agents that behave similarly to human players. Furthermore, we find that input entropy and information gain are similar objectives while empowerment may offer complementary benefits, and thus recommend future work on combining intrinsic objectives.

\paragraph{Future work}

To assign the agent observations to buckets, we naively downscaled them, which is simple but does not account for the semantic similarity between images. We suggest learning the representations using deep neural networks as a direction for future work. A limitation of the human similarity values is that the human datasets are relatively small and it is unclear what instructions the human players received, which could affect how much the players focus on apparent tasks compared to open-ended behavior. Access to more human data and control over the instructions given to the players would be helpful for future work.

%% file: sections/appendix.tex
\onecolumn

\section{Objectives}
\label{sec:metrics}
\input{tables/metrics}
\clearpage

\section{Correlations}
\label{sec:corr}
\input{tables/corr}
\clearpage

\section{Scatter Plots}
\label{fig:scatter}
\input{figures/scatter/figure}
\clearpage

\section{Information Gain Variants}
\label{sec:infogain_variants}

\begin{table}[h!]
\centering
\begin{tabularx}{0.9\textwidth}{Xcccc}
\toprule
& \multicolumn{2}{c}{Shared Discretizations} & \multicolumn{2}{c}{Unshared Discretizations} \\
\textbf{Information Gain Variant} & \textbf{Task Reward} & \textbf{Human Sim.} & \textbf{Task Reward} & \textbf{Human Sim.} \\
\midrule
Dirichlet of transition counts & $\mathllap{-}0.44$ & $\mathllap{-}0.36$ & $0.05$ & $0.40$ \\
Dirichlet of unique transitions & $0.49$ & $0.79$ & $0.35$ & $0.71$ \\
Logarithm of transition counts & $0.55$ & $0.84$ & $0.37$ & $0.78$ \\
Square root of transition counts & $0.52$ & $0.84$ & $0.38$ & $0.78$ \\
\bottomrule
\end{tabularx}
\caption{Correlations between four information gain implementations and task reward and human similarity. We compare the implementations for two forms of preprocessing, using the same discretization levels across agents or using a different discretization for each agent based on its percentiles only. We find that the Dirichlet distribution of unique transitions and logarithm and square root of transition counts exhibit greater correlations when using shared discretizations, and that all of the said three methods correlate strongly with human similarity and more weakly with task reward.}
\end{table}

\section{Human Similarity Variants}

\begin{table}[h!]
\centering
\begin{tabularx}{0.9\textwidth}{Xcccc}
\toprule
\textbf{Human Similarity Variant} & \textbf{Task Reward} & \textbf{Input Entropy} & \textbf{Information Gain} & \textbf{Empowerment} \\
\midrule
Jaccard Similarity & $0.67$ & $0.89$ & $0.79$ & $0.66$ \\
Jensen-Shannon Divergence & $0.26$ & $0.77$ & $0.67$ & $0.66$ \\
\bottomrule
\end{tabularx}
\caption{Correlations between two human similarity implementations and the other objectives. We compare the Jaccard similarity (intersection over union) of the set of inputs visited by the human player and those visited by the RL agent, with the Jensen-Shannon divergence between the two sets. We find that Jaccard similarity correlates more strongly with task reward, slightly more strongly with input entropy and information gain, and near-equally with empowerment, as compared to Jensen-Shannon divergence. The two implementations have a correlation of 0.78 with each other.}
\end{table}

\clearpage

\twocolumn

\section{Environments}
\label{sec:envs}

We consider four environments: Breakout, Seaquest, Montezuma's Revenge, and Minecraft Treechop. The first three are 2D Atari games with backgrounds fixed relative to the screen. The player is free to move around within the screen on one axis in Breakout, and two axes in the Seaquest and Montezuma. In Montezuma, the player can additionally navigate from one room into another. Minecraft is a 3D block-based game in which the player is free to explore, build, and mine; the Treechop task requires the agent to collect wood from trees to obtain task reward.

\paragraph{Breakout}
Breakout is a game in which the agent controls a paddle at the bottom of the screen with the objective of bouncing a ball between the paddle and the blocks above, which disappear upon contact with the ball. The game is nearly deterministic: the only source of randomness other than sticky actions is the initial direction of the ball after starting the game or losing a life. Breakout is the simplest of our three environments, the player only being free to move a paddle in one dimension.

\paragraph{Seaquest}
Seaquest is a game in which the player controls a submarine, with the objective of defending oneself against sharks and other submarines which appear frequently and randomly at both sides of the screen. Additionally, the agent is tasked with picking up divers, which also appear at random, and bringing them to the top of the screen. Because of the random appearance of sharks and divers, the game can be difficult to predict, and made more so by sticky actions. It is more challenging that Breakout, as the player moves along two dimensions and enemies appear at random; however, the agent's task is reactive, with no long-term planning required.

\paragraph{Montezuma}
Montezuma's Revenge is a difficult platformer game with a large first level consisting of many rooms, which necessitates long-term planning. The player must navigate ladders, ropes, and various hazards such as moving skulls and lasers. Rewards are sparse, and given only when the player completes an objective such as finding a key or opening a door, which often require complex and specific action sequences. For this reason, intrinsic rewards \citep{burda2018rnd} or human demonstrations \citep{aytar2018playing} have been important to succeed at the game.

\paragraph{Minecraft Treechop}
MineRL \citep{guss2019minerldata} is a set of environments in Minecraft, a block-based 3D game in which the player can explore, build, and mine within a procedurally generated world. The Treechop environment provides the player with an axe and restricts the action space such that the player can walk around and break blocks, being given task reward for breaking and collecting wood blocks from trees. Though the goal is clearly defined, there are a wide range of possible activities the agent can pursue.

We follow the standard Atari protocol \citep{machado2018stickyactions}. Atari games yield $210 \times 160 \times 3$ images, which we converte to grayscale and rescale to $84 \times 84$ for the agents; MineRL uses $64 \times 64$ images which are input to the agent directly. The agent chooses one of a set of discrete actions: 4 in Breakout, 18 in Seaquest and Montezuma, and 10 in Minecraft. While the effects of the actions in the original games are deterministic, we use ``sticky actions'', meaning that the agent's action is ignored with $25\%$ chance and instead the previous action is repeated.

\section{Agents}
\label{sec:agents}

We selected seven agents to span a wide range of behaviors: no-op and random agents cover low-entropy and high-entropy extremes and the RL agents maximize task reward, intrinsic objectives, and combinations thereof. We use the following agent implementations:

{\raggedright
\begin{itemize}
\item ICM: \href{https://github.com/openai/large-scale-curiosity}{\texttt{https://github.com/openai/ \newline large-scale-curiosity}}
\item RND: \href{https://github.com/openai/random-network-distillation}{\texttt{https://github.com/openai/ \newline random-network-distillation}}
\item PPO: \href{https://github.com/hill-a/stable-baselines}{\texttt{https://github.com/hill-a/ \newline stable-baselines}}
\end{itemize}}

\paragraph{Random}
An agent that uniformly samples random actions from the available action space.
graph{No-op}
The three environments we consider have a no-op action which does nothing, though the environment still continues to update. We consider an agent which always takes this no-op action.

\paragraph{PPO}
Proximal Policy Optimization \citep{schulman2017ppo} trained on extrinsic rewards only. PPO is a commonly used policy gradient algorithm that optimizes the task reward on-policy. It optimizes task reward while preventing the policy from changing too quickly during each training step, stabilizing the learning process.

\paragraph{ICM}
Intrinsic Curiosity Module \citep{pathak2017icm} is an exploration agent seeks inputs that incur high prediction error of a forward model in a learned feature space. Image embeddings that focus on information relevant to the agent are learned by predicting the action from the current and next input. A second network is trained to predict the next embedding from the current embedding an action, and its prediction error serves as the intrinsic reward.

\paragraph{RND}
Random Network Distillation \citep{burda2018rnd} is an exploration agent that seeks inputs that incur high prediction error of a model that predicts image embeddings. The image embeddings are generated by a randomly initialized and fixed neural network. The prediction error serves as the intrinsic reward.

%% file: tables/metrics.tex
\begin{table}[bh!]
\centering
\begin{tabularx}{0.9\textwidth}{clrrrrrrr}
\multicolumn{9}{c}{\textbf{Breakout}} \\
\toprule
Symbol & Objective & No-op & Random & ICM$_\mathrm{n}$ & RND$_\mathrm{n}$ & PPO & ICM$_\mathrm{r}$ & RND$_\mathrm{r}$ \\
\midrule
R & Task Reward & $0.0000$ & $0.0071$ & $0.0695$ & $0.0786$ & $0.0576$ & $\mathbf{0.1302}$ & $0.0743$ \\
S & Human Similarity & $0.0000$ & $0.0247$ & $\mathbf{0.0346}$ & $0.0346$ & $0.0317$ & $0.0346$ & $0.0263$ \\
C & Input Entropy & $0.0000$ & $7.9303$ & $16.2240$ & $15.9964$ & $15.0695$ & $\mathbf{16.6389}$ & $14.9995$ \\
I & Information Gain & $0.0000$ & $0.0203$ & $0.3458$ & $0.3213$ & $0.2770$ & $\mathbf{0.3533}$ & $0.2596$ \\
E & Empowerment & $0.0000$ & $\mathbf{0.4039}$ & $0.4028$ & $0.3907$ & $0.3479$ & $0.3548$ & $0.2770$ \\
\bottomrule
\\[-2ex]
\multicolumn{9}{c}{\textbf{Seaquest}} \\
\toprule
Symbol & Objective & No-op & Random & ICM$_\mathrm{n}$ & RND$_\mathrm{n}$ & PPO & ICM$_\mathrm{r}$ & RND$_\mathrm{r}$ \\
\midrule
R & Task Reward & $0.0000$ & $0.1546$ & $0.1789$ & $0.1424$ & $0.7638$ & $\mathbf{1.7379}$ & $1.0993$ \\
S & Human Similarity & $0.0001$ & $0.0031$ & $0.0034$ & $0.0035$ & $0.0015$ & $\mathbf{0.0037}$ & $0.0032$ \\
C & Input Entropy & $6.7031$ & $12.4245$ & $16.6600$ & $\mathbf{17.0014}$ & $12.6777$ & $14.6161$ & $15.7433$ \\
I & Information Gain & $0.0000$ & $0.2209$ & $0.3700$ & $\mathbf{0.3835}$ & $0.1941$ & $0.2779$ & $0.3319$ \\
E & Empowerment & $0.0000$ & $0.6110$ & $0.4013$ & $0.4008$ & $\mathbf{0.8661}$ & $0.6806$ & $0.5975$ \\
\bottomrule
\\[-2ex]
\multicolumn{9}{c}{\textbf{Montezuma}} \\
\toprule
Symbol & Objective & No-op & Random & ICM$_\mathrm{n}$ & RND$_\mathrm{n}$ & PPO & ICM$_\mathrm{r}$ & RND$_\mathrm{r}$ \\
\midrule
R & Task Reward & $0.0000$ & $0.0003$ & $0.0000$ & $0.2163$ & $0.0003$ & $1.0620$ & $\mathbf{4.2374}$ \\
S & Human Similarity & $0.0001$ & $0.0069$ & $0.0063$ & $0.0081$ & $0.0070$ & $0.0075$ & $\mathbf{0.0120}$ \\
C & Input Entropy & $1.9008$ & $7.1812$ & $7.9123$ & $\mathbf{8.4405}$ & $7.1757$ & $7.1073$ & $7.4942$ \\
I & Information Gain & $0.0000$ & $0.0120$ & $0.0165$ & $\mathbf{0.0229}$ & $0.0113$ & $0.0126$ & $0.0125$ \\
E & Empowerment & $0.0000$ & $0.1326$ & $\mathbf{0.2629}$ & $0.1743$ & $0.1436$ & $0.1869$ & $0.1373$ \\
\bottomrule
\\[-2ex]
\multicolumn{9}{c}{\textbf{Minecraft}} \\
\toprule
Symbol & Objective & No-op & Random & ICM$_\mathrm{n}$ & RND$_\mathrm{n}$ & PPO & ICM$_\mathrm{r}$ & RND$_\mathrm{r}$ \\
\midrule
R & Task Reward & $0.0000$ & $0.0012$ & --- & $\mathbf{0.0013}$ & $0.0012$ & --- & $0.0011$ \\
S & Human Similarity & $0.000002$ & $0.00003$ & --- & $\mathbf{0.00004}$ & $0.00003$ & --- & $0.00003$ \\
C & Input Entropy & $9.4112$ & $\mathbf{16.2270}$ & --- & $14.8039$ & $15.3758$ & --- & $15.0612$ \\
I & Information Gain & $0.0004$ & $\mathbf{0.0583}$ & --- & $0.0345$ & $0.0502$ & --- & $0.0477$ \\
E & Empowerment & $0.0000$ & $0.0770$ & --- & $0.2885$ & $0.1077$ & --- & $\mathbf{0.3444}$ \\
\bottomrule
\end{tabularx}\hfil
\caption{Lifetime values of each objective for all agents and environments, with the highest value of each row in bold. In all three environments, the highest task reward is achieved by task-specific RND or ICM, which maximize both task reward and different implementations of input entropy; and the highest input entropy is achieved by task-agnostic RND or ICM, which is to be expected as these agents use a form of input entropy as their objective. Agents with the highest human similarity, empowerment, and information gain vary by environment. The agents achieve low reward in Minecraft because we only simulated 12M frames due to the slow simulator. We conjecture that the random agent achieves high input entropy and information gain in Minecraft because moving the camera randomly results in many visually different inputs.}
\label{tab:metrics}
\end{table}
 

%% file: tables/corr.tex
\begin{table}[bh!]
\centering
\begin{tabularx}{0.9\textwidth}{lccccc}
\multicolumn{6}{c}{\textbf{Breakout}} \\
\toprule
& Task Reward & Human Similarity & Input Entropy & Information Gain & Empowerment \\
\midrule
Task Reward & 1.00 & 0.74 & 0.85 & 0.91 & 0.48 \\
Human Similarity & 0.74 & 1.00 & 0.96 & 0.83 & 0.93 \\
Input Entropy & 0.85 & 0.96 & 1.00 & 0.94 & 0.79 \\
Information Gain & 0.91 & 0.83 & 0.94 & 1.00 & 0.58 \\
Empowerment & 0.48 & 0.93 & 0.79 & 0.58 & 1.00 \\
\bottomrule
\\[-2ex]
\multicolumn{6}{c}{\textbf{Seaquest}} \\
\toprule
& Task Reward & Human Similarity & Input Entropy & Information Gain & Empowerment \\
\midrule
Task Reward & 1.00 & 0.37 & 0.28 & 0.22 & 0.61 \\
Human Similarity & 0.37 & 1.00 & 0.89 & 0.90 & 0.44 \\
Input Entropy & 0.28 & 0.89 & 1.00 & 1.00 & 0.46 \\
Information Gain & 0.22 & 0.90 & 1.00 & 1.00 & 0.40 \\
Empowerment & 0.61 & 0.44 & 0.46 & 0.40 & 1.00 \\
\bottomrule
\\[-2ex]
\multicolumn{6}{c}{\textbf{Montezuma}} \\
\toprule
& Task Reward & Human Similarity & Input Entropy & Information Gain & Empowerment \\
\midrule
Task Reward & 1.00 & 0.69 & 0.19 & 0.03 & 0.00 \\
Human Similarity & 0.69 & 1.00 & 0.83 & 0.65 & 0.56 \\
Input Entropy & 0.19 & 0.83 & 1.00 & 0.91 & 0.86 \\
Information Gain & 0.03 & 0.65 & 0.91 & 1.00 & 0.80 \\
Empowerment & 0.00 & 0.56 & 0.86 & 0.80 & 1.00 \\
\bottomrule
\\[-2ex]
\multicolumn{6}{c}{\textbf{Minecraft}} \\
\toprule
& Task Reward & Human Similarity & Input Entropy & Information Gain & Empowerment \\
\midrule
Task Reward & 1.00 & 0.96 & 0.98 & 0.91 & 0.59 \\
Human Similarity & 0.96 & 1.00 & 0.89 & 0.78 & 0.74 \\
Input Entropy & 0.98 & 0.89 & 1.00 & 0.98 & 0.49 \\
Information Gain & 0.91 & 0.78 & 0.98 & 1.00 & 0.38 \\
Empowerment & 0.59 & 0.74 & 0.49 & 0.38 & 1.00 \\
\bottomrule
\\[-2ex]
\multicolumn{6}{c}{\textbf{Combined}} \\
\toprule
& Task Reward & Human Similarity & Input Entropy & Information Gain & Empowerment \\
\midrule
Task Reward & 1.00 & 0.67 & 0.54 & 0.49 & 0.41 \\
Human Similarity & 0.67 & 1.00 & 0.89 & 0.79 & 0.66 \\
Input Entropy & 0.54 & 0.89 & 1.00 & 0.95 & 0.66 \\
Information Gain & 0.49 & 0.79 & 0.95 & 1.00 & 0.55 \\
Empowerment & 0.41 & 0.66 & 0.66 & 0.55 & 1.00 \\
\bottomrule
\end{tabularx}%
\caption{Pearson correlation coefficients for each environment and combined; see \cref{fig:corr}.}
\label{tab:corr}
\end{table}

%% file: figures/scatter/figure.tex
\begin{figure}[h!]
\centering
\includegraphics[width=0.8\textwidth]{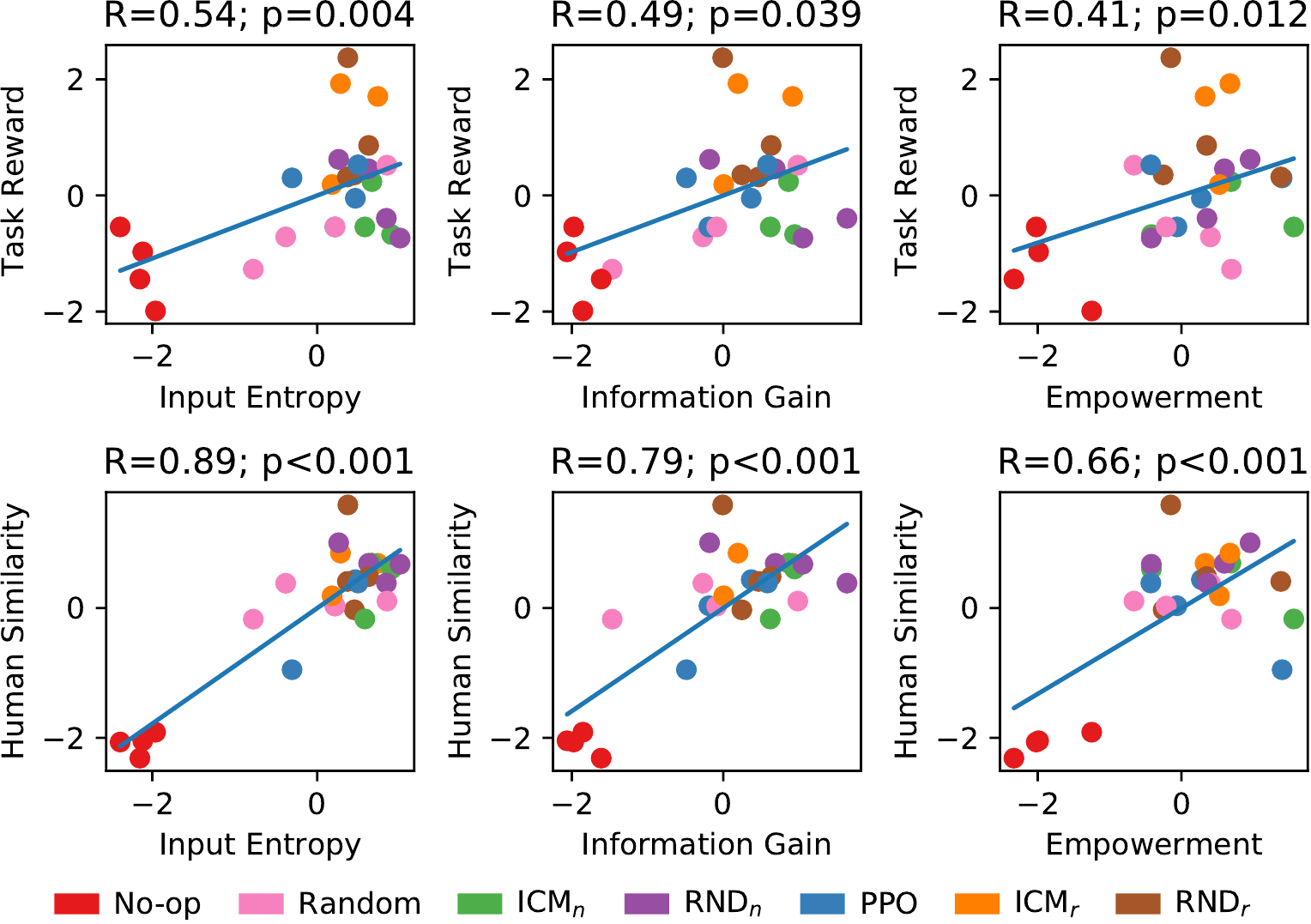}
\caption{Normalized objective values for all agents and environments. The plots visualize the correlations between three intrinsic objectives (X axis) and the two supervised objectives (Y axis). The intrinsic objectives show large correlations with human similarity but their correlations with task reward are weaker. The intrinsic objectives thus capture aspects of agent behavior that is not captured well by the task rewards. For example, they distinguish the qualitatively different behavior of no-op and random agents, even though they obtain similar task reward.}
\end{figure}